%% file: bmvc_review.tex
\documentclass{bmvc2k}

\title{Challenging the Universal Representation of Deep Models for 3D Point Cloud Registration}

\addauthor{David Bojanić}{david.bojanic@fer.hr}{1}
\addauthor{Kristijan Bartol}{kristijan.bartol@fer.hr}{1}
\addauthor{Josep Forest}{ josep.forest@udg.edu}{2}
\addauthor{Stefan Gumhold}{stefan.gumhold@tu-dresden.de}{3}
\addauthor{Tomislav Petković}{tomislav.petkovic.jr@.hr}{1}
\addauthor{Tomislav Pribanić}{tomislav.pribanic@.hr}{1}

\addinstitution{
 Faculty of Electrical Engineering and Computing, University of Zagreb \\
 Zagreb, Croatia
}
\addinstitution{
 Computer Vision and Robotics Research Institute (VICOROB) \\
 University of Girona, Girona, Spain
}
\addinstitution{
TU Dresden, Dresden, Germany
}

\runninghead{Bojanić \etal}{Challenging Universal Representations}

\def\etal{\emph{et al}\bmvaOneDot}

\usepackage{bbm}
\usepackage{multirow}
\usepackage{array}
\usepackage{amsmath}
\usepackage{amsfonts}
\usepackage{eucal}
\DeclareMathOperator*{\argmax}{\arg\!\max}
\usepackage{color, colortbl}
\definecolor{LightCyan}{rgb}{0.88,1,1}

\begin{document}

\maketitle

\begin{abstract}
Learning universal representations across different applications domain is an open research problem. In fact, finding universal architecture within the same application but across different types of datasets is still unsolved problem too, especially in applications involving processing 3D point clouds. In this work we experimentally test several state-of-the-art learning-based methods for 3D point cloud registration against the proposed non-learning baseline registration method. The proposed method either outperforms or achieves comparable results w.r.t. learning based methods. In addition, we propose a dataset on which learning based methods have a hard time to generalize.  Our proposed method and dataset, along with the provided experiments, can be used in further research in studying effective solutions for universal representations. Our source code is available at: \url{github.com/DavidBoja/greedy-grid-search}.
\end{abstract}


\section{Introduction}

3D point cloud registration is the problem of finding an optimal rotation and translation that aligns two overlapping 3D point clouds. 
It arises as a subtask in many different computer vision applications such as: 3D reconstruction \cite{3Dreconstruction_reference1,3Dreconstruction_reference2,3Dreconstruction_reference3}, object recognition and categorization \cite{SHOT,RoPS,object_recognition_reference}, shape retrieval \cite{shape_retrieval_reference}, robot navigation \cite{robot_navigation_reference1} and is still a very researched area \cite{comprehensive-survey,3d-registration-review}.

The most recent advances have been inspired by the successes of deep learning, i.e., by the development of novel architectures and layers convenient for point cloud processing, such as PointNet \cite{pointnet} and 3D convolutions \cite{3d-convolutional-network-for-human-action, c3d-3d-convolutional-networks}. Most of the learning-based approaches first extract point cloud features \cite{dcp-learning-representation-for-registration, prnet-self-supervised--for-partial-to-partial-registration, robust-point-cloud-registration-deep-graph-matching, pcam-product-of-cross-attention-matrices}, and then either apply RANSAC for matching \cite{3dmatch, predator, d3feat-joint-learning-of-dense-features} or learn the whole registration pipeline end-to-end \cite{pointnetlk-robust-and-efficient-registration-using-pointnet, pointnetlk-revisited, omnet-learning-overlapping-mask}. Current state-of-the-art methods \cite{predator, geometric-transformer-for-fast-registration, REGTR, SpinNet, DIP, YOHO, d3feat-joint-learning-of-dense-features} achieve remarkable performance on public benchmarks \cite{3dmatch, KITTI, ETH, modelnet40}, even on very difficult examples with an overlap smaller than 30\% percent \cite{predator}.

The main limitation of the state-of-the-art methods, which is typical for deep-learning based methods \cite{exploring-generalization-in-deep-learning, generalization-in-deep-learning}, is that model performance drops on datasets that differ from the training data. Several recent methods \cite{SpinNet, DIP, geometric-transformer-for-fast-registration, PointDSC} address the generalizability issue and demonstrate significant performance retention between 3DMatch \cite{3dmatch}, KITTI \cite{KITTI}, and ETH \cite{ETH} datasets. To further test the generalization of the existing learning-based methods, we propose a novel FAUST-partial evaluation benchmark. The benchmark consists of overlapping partial views (>60\%), based on the FAUST dataset \cite{FAUST} of 3D human scans, and is substantially different from other public benchmarks. We analyze the performance of the best methods under greater scrutiny by using common 3D registration performance metrics and several distance and angle thresholds.

The aim of this paper is to propose a simple, straightforward 3D registration method and use it as a baseline for comparison with the state-of-the-art, particularly to evaluate their generalization performance. We show competitive performance on KITTI and ETH datasets, when we compare against the models pretrained on 3DMatch. When we compare on a substantially different, FAUST-partial benchmark, we outperform state-of-the-art. This result suggests that the learning-based methods, although remarkable on many public datasets, are still not robust enough to be applied on any 3D data. On the other hand, our baseline method performs consistently, regardless of the data distribution. 

The proposed method is based on the step-wise (grid) search over the possible rotations and translations. 
The point clouds are firstly voxelized. Then, the best transformation candidate is selected as the solution that has the maximum cross-correlation between the two voxel volumes; thus, we call our method \textit{greedy} grid search.

In summary, we:
\begin{itemize}
    \item Evaluate the generalization performance of the state-of-the-art methods under a common set of 3D registration metrics; 
    
    \item Generate a specialized benchmark, called FAUST-partial, based on 3D human scans, which further challenges the generalization of learning-based methods.

    \item Propose a novel 3D registration baseline which selects the transformation candidate based on the maximum cross-correlation between the voxelized point clouds;
    
    \item Demonstrate comparable performance to the state-of-the-art 3D registration methods on public benchmarks and outperform them on FAUST-partial.
\end{itemize}

\section{Related Work}

\textbf{Classical registration.} The most popular classical registration method is the iterative closest point (ICP) algorithm. The algorithm selects a subset of points based on a criteria, calculates the optimal transformation between the clouds using SVD, and iterates until convergence. The original implementations used point-to-point \cite{point2point-icp} and point-to-plane \cite{point2plane-icp} distances for finding the solution, but many other strategies have been proposed \cite{aa-icp, go-icp, trimmed-icp, iterative-global-similarity-points}. GO-ICP \cite{go-icp} proposes a branch-and-bound scheme and proves the global optimality of the algorithm. The 4-point congruent sets (4PCS) algorithm \cite{4pcs} and its variants \cite{v4pcs, generalized-4pcs} are based on the idea that there exist sets of four coplanar points whose alignment corresponds to the alignment of the point clouds. To select the correspondences, RANSAC is used, and ICP is applied for refinement.

\textbf{Handcrafted features.} Methods based on handcrafted features first extract correspondences between the point clouds and then find the transformation using RANSAC. Similar to the image keypoint-based methods such as SIFT \cite{sift}, 3D feature-based methods focus on keypoint detection \cite{a-concise-and-provably-informative-heat-diffusion, harris-3d, learning-a-descriptor-specific-keypoint-detector} and their distinctive description \cite{narf-3d-range-features, fast-point-feature-histograms, unique-shape-context-for-3d-data-description, intrinsic-shape-signatures, using-spin-images-for-object-recognition, recognizing-objects-in-range-data, unique-signatures-of-shistograms, 3d-free-form-object-recognition, keypoint-based-4-pcs}. A few recent methods \cite{cofinet-reliable-coarse-to-fine-correspondences, geometric-transformer-for-fast-registration} retrieve correspondences without keypoint detection by considering all possible matches.

\textbf{Feature learning.} Instead of handcrafting distinctive features, keypoint detection and description can be learned. 3DMatch \cite{3dmatch} transforms patches into volumetric voxel grids of truncated distance function (TDF) values
and processes them through a 3D convolutional network \cite{c3d-3d-convolutional-networks, 3d-convolutional-network-for-human-action} to output local descriptors. Followed by 3DMatch and the popularity of deep learning, many other works propose to learn keypoint detection \cite{d3feat-joint-learning-of-dense-features, usip-unsupervised-stable-interest-point-detection, 3dfeat-net-weakly-supervised} and description \cite{fully-convolutional-geometric-features, ppfnet-global-context-local-features, learning-compact-geometric-features, predator}. Most of these works are learned by optimizing some version of contrastive loss \cite{contrastive-loss1, contrastive-loss2} between the descriptors of matching and non-matching points, and then by applying RANSAC to select the final correspondences.

\textbf{End-to-end registration learning.} There are many recent approaches that learn not only feature description, but also the subsequent matching step, thus learning end-to-end. The first group of these methods \cite{robust-point-cloud-registration-deep-graph-matching, prnet-self-supervised--for-partial-to-partial-registration, dcp-learning-representation-for-registration, rpm-net-robust-matching-using-learned-features, sticky-pillars, iterative-distance-aware-similarity-matrix, pcam-product-of-cross-attention-matrices, deep-gmr}, pioneered by the deep closest point \cite{dcp-learning-representation-for-registration}, follow the ICP idea by iteratively establishing soft correspondences and then applying weighted SVD to obtain the transformations. The remaining group of methods \cite{pointnetlk-robust-and-efficient-registration-using-pointnet, pointnetlk-revisited, feature-metric-registration-without-correspondences, omnet-learning-overlapping-mask, pcrnet}, represented by PointNetLK \cite{pointnetlk-robust-and-efficient-registration-using-pointnet} use PointNet architecture \cite{pointnet} or similar global description strategy to regress the transformation based on the global feature vectors.

\textbf{Generalization to other datasets.} Several recent methods \cite{SpinNet, DIP, PointDSC} attempt to generalize to datasets other than training. All of these methods demonstrate significant performance retention on novel datasets, for example, when evaluating 3DMatch-pretrained models on KITTI \cite{SpinNet, PointDSC}. On the other hand, the results reported on the ETH dataset only show that the computed descriptors have a high recall \cite{SpinNet,d3feat-joint-learning-of-dense-features,DIP} (using the feature-matching-recall measurement), but never actually evaluate the 3D registration.

\begin{figure*}[t!]
\begin{center}
\includegraphics[width=\linewidth]{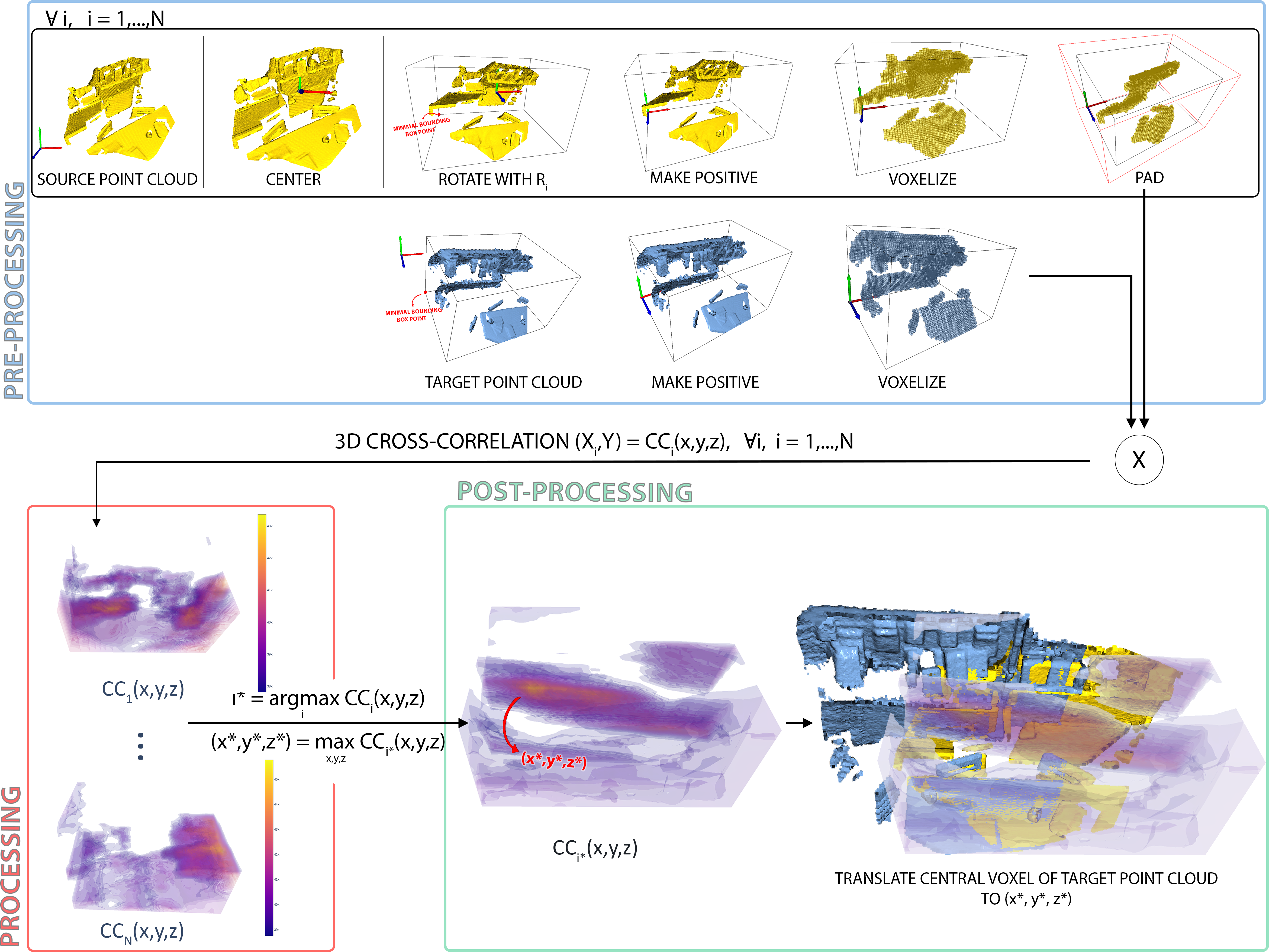}
\end{center}
   \caption{The proposed pipeline. The method is divided into 3 steps: \textit{pre-processing}, \textit{processing} and \textit{post-processing}. Each step follows the previous one. The \textit{pre-processing} step prepares the initial data and outputs $N$ voxelized source volumes and one target volume. The \textit{processing} step performs the 3D cross-correlation over each source volume and the target volume. The cross-correlation volumes $CC_i(x,y,z)$ are heatmaps that should indicate higher (indicated in yellow on the volumes) or lower (indicated with purple on the volumes) matching between the source $\mathbf{X}_i$ and target $\mathbf{Y}$ volumes at the corresponding voxels. White spaces are present because the cross-correlations values are clipped so only the upper range is visible. Finally, the \textit{post-processing} step finds the solution from the output volumes by finding the maximal cross-correlation from all the given volumes.}
\label{fig:pipeline}
\end{figure*}

\section{Method description} \label{sec:method-description}

Let $\mathcal{X} \in \mathbbm{R}^{N \times 3}$ be the \textit{source} point cloud and $\mathcal{Y} \in \mathbbm{R}^{M \times 3}$ the \textit{target} point cloud. The goal of 3D registration is to find the rigid homogeneous transformation $\mathbf{T} \in \text{SE}(3)$ that best aligns $\mathcal{X}$ to $\mathcal{Y}$. The rigid transformation $\mathbf{T}$ is composed of a rotation component $\mathbf{R} \in \text{SO}(3)$ and a translation component $\mathbf{t} \in \mathbbm{R}^3$.

The method pipeline is summarized in Fig. \ref{fig:pipeline}. We divide our method in 3 key steps: \textit{pre-processing}, \textit{processing} and \textit{post-processing}. In this Section we present the general pipeline. The final estimation of the rigid transformation is provided in Eq. \ref{eq:summarized_pipeline2}.

\textbf{R and t parametrization.} To find the correct rotation and translation, we perform a grid search over the rotation and translation space. The rotation space is sampled using 3 Euler angles $\alpha, \beta \text{ and } \gamma$ that rotate the source point cloud $\mathcal{X}$ around the $x,y \text{ and } z$ axes respectively. In a typical 3D scanning use case, two scanned point clouds that need to be registered should be fairly close in the rotation space since the scanning process is limited by a required overlap region. 
We uniformly sample each angle from the range $\left[ -\gamma, \gamma \right]$ with an angle step of $S$ and create a Cartesian product of all the possible combinations resulting in $N$ triplets. Finally, we convert these to rotation matrices $R_i, i=1,\dots, N$. Note that this step is only computed once, prior to registration.
The translation space is inherently sampled by the voxelization process of the given point clouds. The possible translations hence correspond to the centers of the source point cloud voxels and are therefore dependent on the voxelization resolution. More details are provided in the following $4$ sections.

\textbf{Pre-processing.} \label{sec:pre-processing} First, we center and rotate the source point cloud $\mathcal{X}$ around the origin using the precomputed rotation matrices $R_i$, obtaining $\mathcal{X}_i, i=1,\dots,N$. Next, we make all the point clouds coordinates positive by translating their minimal bounding box point into the origin. This step is only done to facilitate the voxelization process and can be completely omitted. We then voxelize each source $\mathcal{X}_i$ and target $\mathcal{Y}$ point clouds with a voxel resolution of $VR$ cm. 
Instead of having a 3D grid with ones and zeros, we set a value of $PV$ (positive voxel) for the filled voxels (voxels where a point from the point cloud is present) and a value of $NV$ (negative voxel) for the empty ones (voxels where there are no points present from the point cloud). 
This results in $N$ source volumes $\mathbf{X}_i$ and one target volume $\mathbf{Y}$. 

\textbf{Processing.} \label{sec:processing} For each source volume $\mathbf{X}_i$, we perform a 3D cross-correlation with the target volume $\mathbf{Y}$. Essentially, the central voxel of the target volume is translated over each voxel of the source volume where the cross-correlation can be computed by multiplying the overlying voxel values of the two volumes and summing them together. This results in $N$ cross-correlation volumes $CC_i(x,y,z)$ with the same 3 dimensions as the source volume. The volumes can be thought of as discrete heatmaps where higher values should represent higher degrees of matching between the voxelized point clouds.
Prior to the cross-correlation, each source volume is padded in order for the target volume to \textit{slide} all over the source volume. We mark with $\mathbf{P} \in \mathbb{R}^6$ the padding applied to each source volume $\mathbf{X}_i$, where the values represent the number of voxels padded to the left, right, top, bottom, front and back of the volume, respectively. 
We make use of the Fourier domain to accelerate the computation of the cross-correlation. Both volumes are first transformed into the Fourier space using the FFT algorithm \cite{fft}, after which the cross-correlation simplifies to a matrix multiplication \cite{fft-book}. The output is then transformed back with an inverse FFT. 

\textbf{Post-processing.} \label{sec:post-processing} We finally estimate the rotation matrix $\hat{R}$ that aligns (rotation-wise) $\mathcal{X}$ to $\mathcal{Y}$ using one of the $N$ precomputed rotation matrices $R_i$. We select the matrix $R_i$ that corresponds to $\mathbf{X}_i$ with the maximal cross-correlation value from the $CC_i(x,y,z)$ volumes. More concretely, we use the
\begin{equation}
    i^* = \argmax_{i} CC_i(x,y,z)
\end{equation}
index to find the estimated rotation matrix $\hat{R} = R_{i^*}$.
To estimate the translation we find the voxel with the maximal cross-correlation value from $CC_{i^*}$. Then, we translate the central voxel of the target volume to the just found voxel of the $CC_{i^*}$ volume. Since the $CC_{i^*}$ volume corresponds to the source $\mathbf{X}_{i^*}$ volume, we essentially translate the central voxel of the target volume to the voxel of the source volume with the maximal cross-correlation. More concretely, we find the index of the voxel with the maximal cross-correlation value with
\begin{equation}
    (x^*,y^*,z^*) = \argmax_{x,y,z} CC_{i^*} (x,y,z).
\end{equation}
Then, to translate the central voxel of the target volume to it, we use the translation:
\begin{equation}
    t_{\text{est}} = \Bigg( \underbrace{- \underbrace{ \mathbf{Y}_{\text{ \tiny CV}}}_{\substack{\text{target} \\ \text{volume} \\ \text{central} \\ \text{voxel} \\ \text{}}}}_{\substack{\text{move to} \\ \text{origin}}} - 
    \underbrace{\underbrace{ \begin{bmatrix}
           \mathbf{P}[0] \\
           \mathbf{P}[2] \\
           \mathbf{P}[4]
         \end{bmatrix}}_{\substack{\text{padding} \\ \text{displacement}}} +
         \underbrace{\begin{bmatrix}
           x^* \\
           y^* \\
           z^*
         \end{bmatrix}}_{\substack{\text{max cc} \\ \text{voxel}}}  + 
          \underbrace{\begin{bmatrix}
           0.5 \\
           0.5 \\
           0.5
         \end{bmatrix}}_{\substack{\text{move to} \\ \text{center of} \\ \text{voxel}}}}_{\text{move to } (x^*,y^*,z^*) } \Bigg) \times \text{VR}
\end{equation}
where each value is multiplied by the voxel resolution $\text{VR}$ to transform from voxel indices to euclidean coordinates. The central voxel of the target volume can be computed as:
\begin{equation}
    \mathbf{Y}_{\text{\tiny CV}} =  \left[ \lceil V_x / 2 \rceil, \lceil V_y / 2 \rceil, \lceil V_z /2 \rceil \right]
\end{equation}
where $V_x,V_y,V_z$ are the number of voxels of $\mathbf{Y}$ along the $3$ dimensions. Intuitively, the central voxel along a dimension is the middle voxel if the number of voxels is odd, and one on the left of the "middle point" if it's odd.

Following all of the steps above, the rigid registration can be summarized as:
\begin{equation} \label{eq:summarized_pipeline1}
     \left( \hat{R} \left( \mathcal{X} - t_{\mathcal{X}}^{\text{ \tiny CENTER}} \right) \right) + t_{\mathcal{X}}^{\text{ \tiny POSIT}} \sim \left( \mathcal{Y} + t_{\mathcal{Y}}^{\text{ \tiny POSIT}} \right) - t_{\text{est}}
\end{equation}
where $\sim$ indicates that the left and right part are aligned. The $t^{\text{ \tiny CENTER}}$ translation moves the center of mass of the respective point cloud into the origin and $t^{\text{ \tiny POSIT}}$ moves the minimal bounding box point into the origin. More concretely:

\begin{equation}
    t_{\mathcal{X}}^{\text{ \tiny CENTER}} = \frac{1}{N} \sum_{i=1}^{N} \mathcal{X}[i,:] \in \mathbb{R}^3, \quad
    t_{\mathcal{X}}^{\text{ \tiny POSIT}} = - \begin{bmatrix}
         \min \mathcal{X}[:,1]  \\
         \min \mathcal{X}[:,2]\\
         \min \mathcal{X}[:,3]
    \end{bmatrix} \in \mathbb{R}^3, \quad
    t_{\mathcal{Y}}^{\text{ \tiny POSIT}} = - \begin{bmatrix}
         \min \mathcal{Y}[:,1]  \\
         \min \mathcal{Y}[:,2]\\
         \min \mathcal{Y}[:,3]
    \end{bmatrix} \in \mathbb{R}^3
\end{equation}

where the $[:,:]$ notation indicates the row-wise and column-wise indexing, and $\min$ indicates the minimal element of an array.

Since the final rigid transformation needs to align $\mathcal{X}$ to $\mathcal{Y}$, Eq. \ref{eq:summarized_pipeline1} equation is further refined as:
\begin{equation} \label{eq:summarized_pipeline2}
    \left( \hat{R} \left( \mathcal{X} - t_{\mathcal{X}}^{\text{ \tiny CENTER}} \right) \right) + t_{\mathcal{X}}^{\text{ \tiny POSIT}} + t_{\text{est}} - t_{\mathcal{Y}}^{\text{ \tiny POSIT}} \sim \mathcal{Y}
\end{equation}

Therefore, the final rotation and translation estimations are:

\begin{equation}
    \hat{R} = R_{i^*},  \hspace{3em} 
    \hat{t} = -\hat{R} \, t_{\mathcal{X}}^{\text{ \tiny CENTER}} + t_{\mathcal{X}}^{\text{ \tiny POSIT}} + t_{\text{est}} - t_{\mathcal{Y}}^{\text{ \tiny POSIT}}
\end{equation}

\textbf{Refinement.} Since the rotation and translation spaces are discretized, the initial alignment is only a rough estimate. The upper bounds on the rotation and translation errors are $\frac{1}{2}\max\limits_{i,j\, i\neq j} \arccos \left( \frac{\text{trace}(R_i^T R_j)-1}{2} \right) \frac{180}{\pi}$ degrees and $\frac{\text{VR} \sqrt{3}}{2}$ centimeters if the ground truth solution is located in the estimated discretized locations. For an angle step of $S=15^\circ$ and $VR=6$cm the upper bound errors would be $7.5^\circ$ and $5$cm. Hence, the rough initial alignment should provide a very good initialization for a fine registration algorithm. 
We use generalized ICP \cite{generalized-icp} to refine the initial solution since it provided the best results.

\section{Experiments} \label{sec:experiments}
We evaluate the generalization capabilities of state-of-the-art methods trained on 3DMatch \cite{3dmatch} and compare them to our baseline. We use two established benchmarking datasets ETH \cite{ETH} and KITTI \cite{KITTI} and create a novel FAUST-partial benchmark based on the FAUST dataset \cite{FAUST}. These datasets test the generalization abilities in terms of different sensor modalities (RGB-D, laser scanner), different environments (indoor, outdoor), resolution ($6$mm to $5$cm) and completely different structure (from indoor objects to humans). 

\textbf{Metrics.} Following \cite{SpinNet,d3feat-joint-learning-of-dense-features,pcam-product-of-cross-attention-matrices,geometric-transformer-for-fast-registration,DGR} we evaluate the results using the Relative Rotation Error (RRE), the Relative Translation Error (RTE) and the Registration Recall (RR) measures. 
The Relative Rotation Error measures the relative angle in degrees between the ground-truth $R^*$ and estimated $\hat{R}$ rotation matrices: 
\begin{equation}
    RRE = \arccos\left(\frac{\text{trace}(\mathbf{\hat{R}}^T \mathbf{R^*}) - 1}{2}\right) \frac{180}{\pi}
\end{equation}
The Relative Translation Error measures the distance from the ground-truth $\mathbf{t^*}$ and estimated $\mathbf{\hat{t}}$ translation vectors:
\begin{equation}
    RTE = \lVert \mathbf{t^*} - \mathbf{\hat{t}}  \lVert_{2}
\end{equation}

The Registration Recall measures the fraction of successfully registered pairs of point clouds. 
A registration is deemed successful (or a true positive in terms of the recall measure) if its RRE and RTE are below predefined thresholds $\tau_r$ and $\tau_t$:

\begin{equation}
    RR = \frac{1}{\lvert \Omega \rvert} \sum_{(i,j) \in \Omega} \mathbbm{1}_{\left\{\text{RRE}(i,j) < \tau_r \hspace{0.2cm} \wedge \hspace{0.2cm} \text{RTE}(i,j) < \tau_t \right\}} 
\end{equation}
where $\Omega$ is the set of all the point cloud registration pairs $(i,j)$ in the dataset, $\mathbbm{1}$ is an indicator function and $\text{RRE}(i,j), \text{RTE}(i,j)$ indicate the $\text{RRE}$ and $\text{RTE}$ for registration pairs $(i,j)$. The final RRE and RTE measurements are averaged only over the successfully registered pairs $(i,j)$ obtained from the RR.

\textbf{Parameters.} To fully define the baseline, we need to set the parameters of the euler angle range $\gamma = 90^\circ$, angle step $S=15^\circ$, the positive voxel value $PV=5$ and the negative voxel value $NV = -1$. The parameters $\gamma$ and $S$ then determine $N=2028$. The padding $\textbf{P}$ is determined for each source volume $\textbf{X}_i$ so that the volume stays the same dimension after the cross-correlation. The only parameter we vary for each dataset is the voxel resolution $VR$ since the datasets vary greatly in their dimensions ranging from volumes of $150\text{m} \times 85\text{m} \times 10\text{m}$ for KITTI to $1.8\text{m} \times 0.8\text{m} \times 0.6\text{m}$ for FAUST-partial.

\textbf{KITTI.} The KITTI dataset \cite{KITTI} is comprised of 11 sequences of outdoor driving scenarios obtained by a LiDAR scanner. Compared to 3DMatch, the fragments are much larger, have lower resolution and a different structure. Following common practice \cite{SpinNet,d3feat-joint-learning-of-dense-features,geometric-transformer-for-fast-registration,predator,fully-convolutional-geometric-features,predator}, we test our baseline on scenes 8 to 10 using pairs which are at least 10m away from each other. The ground truth transformation matrices are refined using ICP and the evaluation thresholds are set to $\tau_r=5^\circ$ and $\tau_t=2$m. We set $VR=75$cm.

As can be seen from Table \ref{tab:KITTI-results}, most state-of-the-art methods tend to generalize well onto the KITTI dataset. The fragments from the dataset are gravity aligned, which is reflected in lower RRE errors since most of the ground truth rotation comes from rotating around one ax. The fragments are also much bigger than those from 3DMatch ($150\text{m} \times 85\text{m} \times 10\text{m}$ on average in volume compared to $3$m$^3$ in 3DMatch) which is reflected in higher RTE errors.
Surprisingly, DIP shows poor recall performance with only $51.71\%$ aligned pairs. The KITTI dataset is much noisier than 3DMatch which might be affecting the local reference frame (LRF) alignment in DIP. 
The baseline achieves comparable results and only lags behind the best recall result from GeoTransformer by less than $1$ percentage point (pp). However, it compensates by achieving the lowest RRE and RTE measurements.

\input{results-KITTI}

\textbf{ETH.} The ETH dataset \cite{ETH} consists of 4 scenes mostly comprised of outdoor vegetation. Compared to 3DMatch, the fragments are larger, have lower resolution and have more complex geometries. Following common practice \cite{the-perfect-match, DIP, SpinNet} we use only point clouds with an overlap greater than $30\%$. We set the thresholds to $\tau_r=5^\circ$ and $\tau_t=30$cm and the voxel resolution of the baseline to $VR=60$cm. The stricter threshold for the RTE reflects the fact that the fragments from ETH are much smaller in volume than those from KITTI.
As can be seen in Table \ref{tab:ETH-results}, the deep learning methods generalize somewhat less than on the KITTI dataset. GeoTransformer and PointDSC achieve low recall results which can be explained by the high RTE error, also visible on the KITTI dataset. Hence, with a stricter RTE threshold, the recall drops significantly. The best performance is achieved by YOHO, followed by the baseline with $6.45$ recall percentage points difference. However, the baseline achieves, again, the lowest RRE and RTE measures between all the methods.

\input{results-ETH}

\textbf{FAUST-partial.} To further test the generalization capabilities of a 3D registration method, we create a new FAUST-partial benchmark based on the FAUST \cite{FAUST} dataset. The state-of-the-art methods that train on the 3DMatch dataset, test their generalization capabilities \cite{SpinNet,DIP,geometric-transformer-for-fast-registration,pcam-product-of-cross-attention-matrices,PointDSC,the-perfect-match} on the KITTI \cite{KITTI} and ETH \cite{ETH} datasets that tend to be comprised of similar objects. We argue that because of that reason, using proper data augmentation whilst learning on 3DMatch, a method can increase its robustness and generalization on noisier and perturbed fragments from other datasets. However, when encountered with completely unseen data (such as 3D human scans), the methods have difficulty generalizing, even when having more than $60\%$ overlap.

\begin{figure*}
\begin{center}

\includegraphics[width=\linewidth]{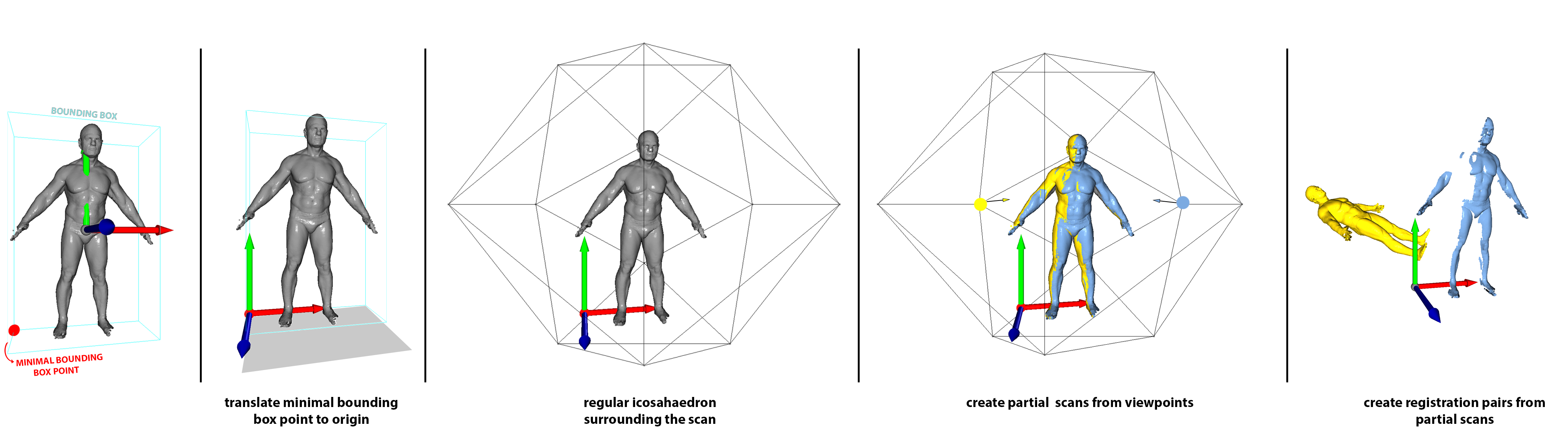}
\end{center} 
   \caption{FAUST-partial dataset generation. For a given scan from the FAUST \cite{FAUST} dataset, we translate its minimal bounding box point to the origin. Next, we surround the scan with a regular icosahaedron. Each point of the icosahaedron acts as a viewpoint used to create a partial scan using the hidden point removal algorithm \cite{hidden_point_removal}. For two partial scans with an overalp bigger than $60\%$, we use a random rotation and translation to obtain a registration pair for the FAUST-partial benchmark.}
\label{fig:faust-partial}
\end{figure*}

To create a 3D registration benchmark, we use the FAUST training dataset comprised of $100$ human body scans. We divide each scan into multiple overlapping fragments that need to be aligned. The steps are illustrated in Fig.~\ref{fig:faust-partial}. To create the fragments, we begin by moving each scan so the $xz$-plane acts as the floor. We do this by moving the minimal bounding box point to the origin. Next, we create a regular icosahaedron centered at the center of mass of each scan. A regular icosahaedron contains $12$ points that lie on a unit sphere around its center, each equidistant from its neighbors. We scale the icosahaedron to a $1.5$m radius sphere so each scan fits inside it. These points are used as the viewpoints for creating the partial views (fragments), discarding the ones located below the $xz$-plane (below the floor). For each viewpoint, we use the hidden point removal \cite{hidden_point_removal} algorithm to create a partial point cloud. 
Next, for each two pairs of viewpoints $(i,j)$ we sample a rotation using 3 Euler angles from the range $\left[ 0^{\circ}, 45^{\circ} \right]$ and a random translation from the range $\left[ -50\text{cm}, 50\text{cm} \right]$. We rotate and translate the partial point cloud obtained from viewpoint $i$ to finally get a benchmark registration pair. For the final benchmark we use every pair of viewpoints that have an overlap bigger than $60\%$, resulting in $1724$ registration pairs.

For evaluating the baseline we set the voxel resolution to $VR=6$cm and use the thresholds $\tau_r=10^\circ$ and $\tau_t=3$cm. The strict threshold for RTE reflects the fact that fragments from FAUST-partial are much smaller in volume than all the other datasets. As can be seen from Table \ref{tab:FAUST-partial-results}, deep learning methods are incapable of generalizing onto datasets that differ considerably from the 3DMatch-like datasets on which they were trained on. Additionally, the FAUST-partial registration pairs have considerably bigger overlaps $(>60\%)$ than the 3DMatch benchmark. However, the best recall of  $56.15\%$ from the deep learning methods is achieved by GeoTransformer. Contrary, the baseline performs consistently, achieving a remarkable $92.81\%$ recall and the lowest RRE and RTE measures.

\input{results-FAUST-partial}

\section{Conclusion}
The proposed classical approach provides a good 3D registration baseline. The method is simple but effective, which is demonstrated on the public benchmarks. Compared to the generalization performance of the state-of-the-art methods, the baseline is on-par. On the newly proposed, FAUST-partial benchmark, the competing methods are struggling to retain the results, or perform significantly worse, even though the generated overlap between the cloud pairs is reasonably high. Contrary to the deep learning methods, the baseline is simple and explainable and serves for detailed analyses. The effects of different strategies are clear and intuitive and provide insights into the registration process. Therefore, on the search of finding universal representations, designing a deep model mimicking the proposed baseline method is an interesting future avenue to pursue. \\
\textbf{Acknowledgement} This work has been supported by the Croatian Science Foundation under the project IP-2018-01-8118.

\bibliography{egbib}
\end{document}

%% file: results-KITTI.tex
\begin{table}[h!]
\scriptsize
\begin{center}
\begin{tabular}{l|c|c|c}
\hline
Method      & RR(\%) & RRE($^\circ$) & RTE (cm) \\
\hline
FCGF$^\dagger$  \cite{fully-convolutional-geometric-features}      & 24.19 & 1.61  & 27.10    \\
D3Feat-rand$^\dagger$ \cite{d3feat-joint-learning-of-dense-features} & 18.47 & 1.58  & 37.80    \\
D3Feat-pred$^\dagger$ \cite{d3feat-joint-learning-of-dense-features}& 36.76 & 1.44  & 31.60    \\
SpinNet$^\dagger$   \cite{SpinNet}  & 81.44 & 0.98  & 15.60    \\
DIP       & 51.71  & 1.02   & 13.43  \\
GeoTransformer \cite{geometric-transformer-for-fast-registration} & $\mathbf{90.63}$  & 0.54 &  154.27 \\
FCGF+PointDSC \cite{PointDSC} & 66.13 & 0.88   & 107.71      \\
FCGF+YOHO-O \cite{YOHO} & 81.44  & 1.99  & 54.25      \\
FCGF+YOHO-C \cite{YOHO} & 82.16  & 1.38   &  39.30     \\
\hline
Baseline  & 90.27 & $\mathbf{0.13}$ & $\mathbf{4.68}$ \\
\hline
\end{tabular}
\caption{Results on the KITTI dataset. All the methods are trained on the 3DMatch dataset. Results marked with $\dagger$ are taken from \cite{SpinNet}.}
\label{tab:KITTI-results}
\end{center}
\end{table}

%% file: results-ETH.tex
\begin{table}[h!]
\scriptsize
\begin{center}
\begin{tabular}{l|c|c|c}
\hline
Method      & RR(\%) & RRE($^\circ$) & RTE (cm) \\
\hline
SpinNet \cite{SpinNet} & 73.07 & 1.205 & 5.352    \\
DIP \cite{DIP} & 62.41 & 1.940 & 14.716    \\
GeoTransformer \cite{geometric-transformer-for-fast-registration} & 4.91 & 0.710  &  21.162    \\
FCGF+PointDSC \cite{PointDSC} &  2.81  &  0.573 &  23.426    \\
FCGF+YOHO-O \cite{YOHO} & 79.94 &  2.167  &  16.112    \\
FCGF+YOHO-C \cite{YOHO} & $\mathbf{84.85}$ & 1.950  &  16.176    \\
\hline
Baseline    &   78.40    &   $\mathbf{0.355}$  &  $\mathbf{1.706}$ \\
\hline
\end{tabular}
\caption{Results on ETH dataset. All the methods are trained on the 3DMatch dataset.}
\label{tab:ETH-results}
\end{center}
\end{table}

%% file: results-FAUST-partial.tex
\begin{table}[t]
\scriptsize
\begin{center}
\begin{tabular}{p{3cm}|
>{\centering\arraybackslash}p{0.65cm} 
>{\centering\arraybackslash}p{0.7cm} 
>{\centering\arraybackslash}p{1.0cm}} 
\hline
Method      & RR(\%)   & RRE($^\circ$)  & RTE(cm)\\
\hline
FPFH-8M \cite{fast-point-feature-histograms} & 9.51 & 4.347  & 1.900  \\
SpinNet \cite{SpinNet} & 42.46 & 3.105 & 1.670  \\
GeoTransformer \cite{geometric-transformer-for-fast-registration} & 56.15 & 2.423  & 1.581  \\
FCGF+PointDSC \cite{PointDSC}  & 47.85  & 3.354  & 1.793 \\
FCGF+YOHO-O \cite{YOHO} & 18.91 & 4.489  & 1.852 \\
FCGF+YOHO-C \cite{YOHO} & 29.18 & 3.653  & 1.668  \\
DIP \cite{DIP} & 54.81 & 4.058  & 2.052  \\
\hline
Baseline    & $\mathbf{92.81}$ & $\mathbf{0.014}$  & $\mathbf{0.009}$  \\
\hline
\end{tabular}
\end{center}
 \caption{Results on FAUST-partial dataset. All the methods are trained on the 3DMatch dataset. FPFH-8M is registered with RANSAC using 8 million iterations.}
 \label{tab:FAUST-partial-results}
\end{table}

%% file: bmvc_review.bbl
\begin{thebibliography}{77}
\providecommand{\natexlab}[1]{#1}
\providecommand{\url}[1]{\texttt{#1}}
\expandafter\ifx\csname urlstyle\endcsname\relax
  \providecommand{\doi}[1]{doi: #1}\else
  \providecommand{\doi}{doi: \begingroup \urlstyle{rm}\Url}\fi

\bibitem[Aiger et~al.(2008)Aiger, Mitra, and Cohen-Or]{4pcs}
Dror Aiger, Niloy~J. Mitra, and Daniel Cohen-Or.
\newblock 4-points congruent sets for robust pairwise surface registration.
\newblock \emph{ACM Trans. Graph.}, 27\penalty0 (3):\penalty0 1–10, aug 2008.
\newblock ISSN 0730-0301.
\newblock \doi{10.1145/1360612.1360684}.
\newblock URL \url{https://doi.org/10.1145/1360612.1360684}.

\bibitem[Ao et~al.(2021)Ao, Hu, Yang, Markham, and Guo]{SpinNet}
Sheng Ao, Qingyong Hu, Bo~Yang, Andrew Markham, and Yulan Guo.
\newblock Spinnet: Learning a general surface descriptor for 3d point cloud
  registration.
\newblock In \emph{Proceedings of the IEEE/CVF Conference on Computer Vision
  and Pattern Recognition}, 2021.

\bibitem[Aoki et~al.(2019)Aoki, Goforth, Srivatsan, and
  Lucey]{pointnetlk-robust-and-efficient-registration-using-pointnet}
Yasuhiro Aoki, Hunter Goforth, Rangaprasad~Arun Srivatsan, and Simon Lucey.
\newblock Pointnetlk: Robust \& efficient point cloud registration using
  pointnet.
\newblock \emph{2019 IEEE/CVF Conference on Computer Vision and Pattern
  Recognition (CVPR)}, pages 7156--7165, 2019.

\bibitem[Bai et~al.(2020)Bai, Luo, Zhou, Fu, Quan, and
  Tai]{d3feat-joint-learning-of-dense-features}
Xuyang Bai, Zixin Luo, Lei Zhou, Hongbo Fu, Long Quan, and Chiew-Lan Tai.
\newblock D3feat: Joint learning of dense detection and description of 3d local
  features.
\newblock \emph{2020 IEEE/CVF Conference on Computer Vision and Pattern
  Recognition (CVPR)}, pages 6358--6366, 2020.

\bibitem[Bai et~al.(2021)Bai, Luo, Zhou, Chen, nad Zeyu~Hu, Fu, and
  Tai]{PointDSC}
Xuyang Bai, Zixin Luo, Lei Zhou, Hongkai Chen, Lei~Li nad Zeyu~Hu, Hongbo Fu,
  and Chiew-Lan Tai.
\newblock {PointDSC}: {R}obust {P}oint {C}loud {R}egistration using {D}eep
  {S}patial {C}onsistency.
\newblock \emph{CVPR}, 2021.

\bibitem[Besl and McKay(1992)]{point2point-icp}
Paul~J. Besl and Neil~D. McKay.
\newblock A method for registration of 3-d shapes.
\newblock \emph{IEEE Trans. Pattern Anal. Mach. Intell.}, 14:\penalty0
  239--256, 1992.

\bibitem[Blais and Levine(1995)]{3Dreconstruction_reference2}
G.~Blais and M.~D. Levine.
\newblock Registering multiview range data to create 3d computer objects.
\newblock \emph{IEEE Transactions on Pattern Analysis and Machine
  Intelligence}, 17\penalty0 (8):\penalty0 820--824, 1995.

\bibitem[Bogo et~al.(2014)Bogo, Romero, Loper, and Black]{FAUST}
Federica Bogo, Javier Romero, Matthew Loper, and Michael~J. Black.
\newblock Faust: Dataset and evaluation for {3D} mesh registration.
\newblock In \emph{Proceedings IEEE Conf. on Computer Vision and Pattern
  Recognition (CVPR)}, Piscataway, NJ, USA, June 2014. IEEE.

\bibitem[Bojani\'{c} et~al.(2020)Bojani\'{c}, Bartol, Petkovi\'{c}, and
  Pribani\'{c}]{3d-registration-review}
David Bojani\'{c}, Kristijan Bartol, Tomislav Petkovi\'{c}, and Tomislav
  Pribani\'{c}.
\newblock A review of rigid 3d registration methods.
\newblock \emph{Proceedings of 13th International Scientific – Professional
  Symposium TEXTILE SCIENCE \& ECONOMY}, 2020.

\bibitem[Brigham and Morrow(1967)]{fft}
E.~O. Brigham and R.~E. Morrow.
\newblock The fast fourier transform.
\newblock \emph{IEEE Spectrum}, 4\penalty0 (12):\penalty0 63--70, 1967.
\newblock \doi{10.1109/MSPEC.1967.5217220}.

\bibitem[Cao et~al.(2021)Cao, Puy, Boulch, and
  Marlet]{pcam-product-of-cross-attention-matrices}
Anh-Quan Cao, Gilles Puy, Alexandre Boulch, and Renaud Marlet.
\newblock {PCAM}: {P}roduct of {C}ross-{A}ttention {M}atrices for {R}igid
  {R}egistration of {P}oint {C}louds.
\newblock In \emph{International Conference on Computer Vision (ICCV)}, 2021.

\bibitem[Chen and Medioni(1992)]{point2plane-icp}
Yang Chen and G{\'e}rard~G. Medioni.
\newblock Object modelling by registration of multiple range images.
\newblock \emph{Image Vis. Comput.}, 10:\penalty0 145--155, 1992.

\bibitem[Chetverikov et~al.(2002)Chetverikov, Svirko, Stepanov, and
  Krsek]{trimmed-icp}
D.~Chetverikov, D.~Svirko, D.~Stepanov, and P.~Krsek.
\newblock The trimmed iterative closest point algorithm.
\newblock In \emph{2002 International Conference on Pattern Recognition},
  volume~3, pages 545--548 vol.3, 2002.
\newblock \doi{10.1109/ICPR.2002.1047997}.

\bibitem[Chopra et~al.(2005)Chopra, Hadsell, and LeCun]{contrastive-loss1}
S.~Chopra, R.~Hadsell, and Y.~LeCun.
\newblock Learning a similarity metric discriminatively, with application to
  face verification.
\newblock In \emph{2005 IEEE Computer Society Conference on Computer Vision and
  Pattern Recognition (CVPR'05)}, volume~1, pages 539--546 vol. 1, 2005.
\newblock \doi{10.1109/CVPR.2005.202}.

\bibitem[Choy et~al.(2019)Choy, Park, and
  Koltun]{fully-convolutional-geometric-features}
Christopher Choy, Jaesik Park, and Vladlen Koltun.
\newblock Fully convolutional geometric features.
\newblock In \emph{2019 IEEE/CVF International Conference on Computer Vision
  (ICCV)}, pages 8957--8965, 2019.
\newblock \doi{10.1109/ICCV.2019.00905}.

\bibitem[Choy et~al.(2020)Choy, Dong, and Koltun]{DGR}
Christopher Choy, Wei Dong, and Vladlen Koltun.
\newblock Deep global registration.
\newblock In \emph{CVPR}, 2020.

\bibitem[Deng et~al.(2018)Deng, Birdal, and
  Ilic]{ppfnet-global-context-local-features}
Haowen Deng, Tolga Birdal, and Slobodan Ilic.
\newblock Ppfnet: Global context aware local features for robust 3d point
  matching.
\newblock In \emph{2018 IEEE/CVF Conference on Computer Vision and Pattern
  Recognition (CVPR)}, 10 2018.
\newblock \doi{10.1109/CVPR.2018.00028}.

\bibitem[Frome et~al.(2004)Frome, Huber, Kolluri, Bülow, and
  Malik]{recognizing-objects-in-range-data}
Andrea Frome, Daniel Huber, Ravi Kolluri, Thomas Bülow, and Jitendra Malik.
\newblock Recognizing objects in range data using regional point descriptors.
\newblock volume~3, pages 224--237, 05 2004.
\newblock ISBN 978-3-540-21982-8.
\newblock \doi{10.1007/978-3-540-24672-5_18}.

\bibitem[Fu et~al.(2021)Fu, Liu, Luo, and
  Wang]{robust-point-cloud-registration-deep-graph-matching}
Kexue Fu, Shaolei Liu, Xiaoyuan Luo, and Manning Wang.
\newblock Robust point cloud registration framework based on deep graph
  matching.
\newblock \emph{2021 IEEE/CVF Conference on Computer Vision and Pattern
  Recognition (CVPR)}, pages 8889--8898, 2021.

\bibitem[Geiger et~al.(2012)Geiger, Lenz, and Urtasun]{KITTI}
Andreas Geiger, Philip Lenz, and Raquel Urtasun.
\newblock Are we ready for autonomous driving? the kitti vision benchmark
  suite.
\newblock In \emph{2012 IEEE Conference on Computer Vision and Pattern
  Recognition}, pages 3354--3361, 2012.
\newblock \doi{10.1109/CVPR.2012.6248074}.

\bibitem[Gojcic et~al.(2019)Gojcic, Zhou, Wegner, and
  Wieser]{the-perfect-match}
Zan Gojcic, Caifa Zhou, Jan~Dirk Wegner, and Andreas Wieser.
\newblock The perfect match: 3d point cloud matching with smoothed densities.
\newblock \emph{2019 IEEE/CVF Conference on Computer Vision and Pattern
  Recognition (CVPR)}, pages 5540--5549, 2019.

\bibitem[Guo et~al.(2013{\natexlab{a}})Guo, Sohel, Bennamoun, Lu, and
  Wan]{RoPS}
Y.~Guo, F.~Sohel, M.~Bennamoun, M.~Lu, and J.~Wan.
\newblock Rotational projection statistics for 3d local surface description and
  object recognition.
\newblock \emph{International Journal of Computer Vision}, 105\penalty0
  (1):\penalty0 63 – 86, Apr 2013{\natexlab{a}}.
\newblock ISSN 1573-1405.
\newblock \doi{10.1007/s11263-013-0627-y}.
\newblock URL \url{http://dx.doi.org/10.1007/s11263-013-0627-y}.

\bibitem[Guo et~al.(2013{\natexlab{b}})Guo, Bennamoun, Sohel, Wan, and
  Lu]{3d-free-form-object-recognition}
Yulan Guo, Mohammed Bennamoun, Ferdous~A. Sohel, Jianwei Wan, and Min Lu.
\newblock 3d free form object recognition using rotational projection
  statistics.
\newblock In \emph{2013 IEEE Workshop on Applications of Computer Vision
  (WACV)}, pages 1--8, 2013{\natexlab{b}}.
\newblock \doi{10.1109/WACV.2013.6474992}.

\bibitem[Huang et~al.(2017)Huang, Kwok, and Zhou]{v4pcs}
Jida Huang, Tsz-Ho Kwok, and Chi Zhou.
\newblock {V4PCS: Volumetric 4PCS Algorithm for Global Registration}.
\newblock \emph{Journal of Mechanical Design}, 139\penalty0 (11), 10 2017.
\newblock ISSN 1050-0472.
\newblock \doi{10.1115/1.4037477}.
\newblock URL \url{https://doi.org/10.1115/1.4037477}.
\newblock 111403.

\bibitem[Huang et~al.(2021{\natexlab{a}})Huang, Gojcic, Usvyatsov, Wieser, and
  Schindler]{predator}
Shengyu Huang, Zan Gojcic, Mikhail~(Misha) Usvyatsov, Andreas Wieser, and
  Konrad Schindler.
\newblock Predator: Registration of 3d point clouds with low overlap.
\newblock \emph{2021 IEEE/CVF Conference on Computer Vision and Pattern
  Recognition (CVPR)}, pages 4265--4274, 2021{\natexlab{a}}.

\bibitem[Huang et~al.(2020)Huang, Mei, and
  Zhang]{feature-metric-registration-without-correspondences}
Xiaoshui Huang, Guofeng Mei, and Jian Zhang.
\newblock Feature-metric registration: A fast semi-supervised approach for
  robust point cloud registration without correspondences.
\newblock \emph{2020 IEEE/CVF Conference on Computer Vision and Pattern
  Recognition (CVPR)}, pages 11363--11371, 2020.

\bibitem[Huang et~al.(2021{\natexlab{b}})Huang, Mei, Zhang, and
  Abbas]{comprehensive-survey}
Xiaoshui Huang, Guofeng Mei, Jian Zhang, and Rana Abbas.
\newblock A comprehensive survey on point cloud registration.
\newblock \emph{ArXiv}, abs/2103.02690, 2021{\natexlab{b}}.

\bibitem[Huber and Hebert(2003)]{3Dreconstruction_reference3}
D.~Huber and M.~Hebert.
\newblock Fully automatic registration of multiple 3d data sets.
\newblock \emph{Image and Vision Computing}, 21:\penalty0 637--650, 07 2003.
\newblock \doi{10.1016/S0262-8856(03)00060-X}.

\bibitem[Ji et~al.(2013)Ji, Xu, Yang, and
  Yu]{3d-convolutional-network-for-human-action}
Shuiwang Ji, Wei Xu, Ming Yang, and Kai Yu.
\newblock 3d convolutional neural networks for human action recognition.
\newblock \emph{IEEE Transactions on Pattern Analysis and Machine
  Intelligence}, 35\penalty0 (1):\penalty0 221--231, 2013.
\newblock \doi{10.1109/TPAMI.2012.59}.

\bibitem[Johnson and Hebert(1999{\natexlab{a}})]{object_recognition_reference}
A.~E. Johnson and M.~Hebert.
\newblock Using spin images for efficient object recognition in cluttered 3d
  scenes.
\newblock \emph{IEEE Transactions on Pattern Analysis and Machine
  Intelligence}, 21\penalty0 (5):\penalty0 433--449, 1999{\natexlab{a}}.

\bibitem[Johnson and
  Hebert(1999{\natexlab{b}})]{using-spin-images-for-object-recognition}
A.E. Johnson and M.~Hebert.
\newblock Using spin images for efficient object recognition in cluttered 3d
  scenes.
\newblock \emph{IEEE Transactions on Pattern Analysis and Machine
  Intelligence}, 21\penalty0 (5):\penalty0 433--449, 1999{\natexlab{b}}.
\newblock \doi{10.1109/34.765655}.

\bibitem[Katz et~al.(2007)Katz, Tal, and Basri]{hidden_point_removal}
Sagi Katz, Ayellet Tal, and Ronen Basri.
\newblock Direct visibility of point sets.
\newblock volume~26, 07 2007.
\newblock \doi{10.1145/1275808.1276407}.

\bibitem[Kawaguchi et~al.(2017)Kawaguchi, Kaelbling, and
  Bengio]{generalization-in-deep-learning}
Kenji Kawaguchi, Leslie~Pack Kaelbling, and Yoshua Bengio.
\newblock Generalization in deep learning.
\newblock \emph{ArXiv}, abs/1710.05468, 2017.

\bibitem[Khoury et~al.(2017)Khoury, Zhou, and
  Koltun]{learning-compact-geometric-features}
Marc Khoury, Qian-Yi Zhou, and Vladlen Koltun.
\newblock Learning compact geometric features.
\newblock \emph{2017 IEEE International Conference on Computer Vision (ICCV)},
  pages 153--161, 2017.

\bibitem[Li et~al.(2020)Li, Zhang, Xu, Zhou, and
  Zhang]{iterative-distance-aware-similarity-matrix}
Jiahao Li, Changhao Zhang, Ziyao Xu, Hangning Zhou, and Chi Zhang.
\newblock Iterative distance-aware similarity matrix convolution with
  mutual-supervised point elimination for efficient point cloud registration.
\newblock In \emph{European Conference on Computer Vision (ECCV)}, 2020.

\bibitem[Li and Lee(2019)]{usip-unsupervised-stable-interest-point-detection}
Jiaxin Li and Gim~Hee Lee.
\newblock Usip: Unsupervised stable interest point detection from 3d point
  clouds.
\newblock \emph{2019 IEEE/CVF International Conference on Computer Vision
  (ICCV)}, pages 361--370, 2019.

\bibitem[Li et~al.(2021)Li, Pontes, and Lucey]{pointnetlk-revisited}
Xueqian Li, Jhony~Kaesemodel Pontes, and Simon Lucey.
\newblock Pointnetlk revisited.
\newblock In \emph{Proceedings of the IEEE/CVF Conference on Computer Vision
  and Pattern Recognition (CVPR)}, pages 12763--12772, June 2021.

\bibitem[Li et~al.(2015)Li, Dai, Guibas, and
  Niessner]{shape_retrieval_reference}
Y.~Li, A.~Dai, L.~Guibas, and M.~Niessner.
\newblock Database-assisted object retrieval for real-time 3d reconstruction.
\newblock \emph{Computer Graphics Forum}, 34:\penalty0 435--446, 2015.

\bibitem[Lowe(1999)]{sift}
David~G Lowe.
\newblock Object recognition from local scale-invariant features.
\newblock In \emph{Proceedings of the seventh IEEE international conference on
  computer vision}, volume~2, pages 1150--1157. Ieee, 1999.

\bibitem[Magnusson et~al.(2007)Magnusson, Lilienthal, and
  Duckett]{robot_navigation_reference1}
M.~Magnusson, A.~Lilienthal, and T.~Duckett.
\newblock Scan registration for autonomous mining vehicles using 3d-ndt.
\newblock \emph{Journal of Field Robotics}, 24:\penalty0 803--827, 10 2007.
\newblock \doi{10.1002/rob.20204}.

\bibitem[Mohamad et~al.(2014)Mohamad, Rappaport, and
  Greenspan]{generalized-4pcs}
Mustafa Mohamad, David Rappaport, and Michael Greenspan.
\newblock Generalized 4-points congruent sets for 3d registration.
\newblock In \emph{2014 2nd International Conference on 3D Vision}, volume~1,
  pages 83--90, 2014.
\newblock \doi{10.1109/3DV.2014.21}.

\bibitem[Neyshabur et~al.(2017)Neyshabur, Bhojanapalli, McAllester, and
  Srebro]{exploring-generalization-in-deep-learning}
Behnam Neyshabur, Srinadh Bhojanapalli, David McAllester, and Nathan Srebro.
\newblock Exploring generalization in deep learning.
\newblock In \emph{NIPS}, 2017.

\bibitem[Pavlov et~al.(2018)Pavlov, Ovchinnikov, Derbyshev, Tsetserukou, and
  Oseledets]{aa-icp}
Artem Pavlov, Grigory Ovchinnikov, Dmitriy Derbyshev, Dzmitry Tsetserukou, and
  Ivan Oseledets.
\newblock Aa-icp: Iterative closest point with anderson acceleration.
\newblock pages 1--6, 05 2018.
\newblock \doi{10.1109/ICRA.2018.8461063}.

\bibitem[Poiesi and Boscaini(2021)]{DIP}
Fabio Poiesi and Davide Boscaini.
\newblock Distinctive {3D} local deep descriptors.
\newblock In \emph{IEEE Proc. of Int'l Conference on Pattern Recognition},
  Milan, IT, Jan 2021.

\bibitem[Pomerleau et~al.(2012)Pomerleau, Liu, Colas, and Siegwart]{ETH}
François Pomerleau, Ming Liu, Francis Colas, and Roland Siegwart.
\newblock Challenging data sets for point cloud registration algorithms.
\newblock \emph{The International Journal of Robotics Research}, 31:\penalty0
  1705--1711, 12 2012.
\newblock \doi{10.1177/0278364912458814}.

\bibitem[Qi et~al.(2017)Qi, Su, Mo, and Guibas]{pointnet}
C.~Qi, Hao Su, Kaichun Mo, and Leonidas~J. Guibas.
\newblock Pointnet: Deep learning on point sets for 3d classification and
  segmentation.
\newblock \emph{2017 IEEE Conference on Computer Vision and Pattern Recognition
  (CVPR)}, pages 77--85, 2017.

\bibitem[Qin et~al.(2022)Qin, Yu, Wang, Guo, Peng, and
  Xu]{geometric-transformer-for-fast-registration}
Zheng Qin, Hao Yu, Changjian Wang, Yulan Guo, Yuxing Peng, and Kaiping Xu.
\newblock Geometric transformer for fast and robust point cloud registration.
\newblock \emph{ArXiv}, abs/2202.06688, 2022.

\bibitem[Rao et~al.(2010)Rao, Kim, and Hwang]{fft-book}
K.~R. Rao, D.~N. Kim, and J.-J. Hwang.
\newblock \emph{Fast Fourier Transform - Algorithms and Applications}.
\newblock Springer Publishing Company, Incorporated, 1st edition, 2010.
\newblock ISBN 1402066287.

\bibitem[Rusu et~al.(2009)Rusu, Blodow, and
  Beetz]{fast-point-feature-histograms}
Radu~Bogdan Rusu, Nico Blodow, and Michael Beetz.
\newblock Fast point feature histograms (fpfh) for 3d registration.
\newblock In \emph{2009 IEEE International Conference on Robotics and
  Automation}, pages 3212--3217, 2009.
\newblock \doi{10.1109/ROBOT.2009.5152473}.

\bibitem[Salti et~al.(2015)Salti, Tombari, Spezialetti, and
  Stefano]{learning-a-descriptor-specific-keypoint-detector}
Samuele Salti, Federico Tombari, Riccardo Spezialetti, and Luigi~Di Stefano.
\newblock Learning a descriptor-specific 3d keypoint detector.
\newblock In \emph{2015 IEEE International Conference on Computer Vision
  (ICCV)}, pages 2318--2326, 2015.
\newblock \doi{10.1109/ICCV.2015.267}.

\bibitem[Sarode et~al.(2019)Sarode, Li, Goforth, Aoki, Srivatsan, Lucey, and
  Choset]{pcrnet}
Vinit Sarode, Xueqian Li, Hunter Goforth, Yasuhiro Aoki, Rangaprasad~Arun
  Srivatsan, Simon Lucey, and Howie Choset.
\newblock {PCRN}et: Point {C}loud {R}egistration {N}etwork using {P}ointnet
  {E}ncoding.
\newblock \emph{ArXiv}, abs/1908.07906, 2019.

\bibitem[Schroff et~al.(2015)Schroff, Kalenichenko, and
  Philbin]{contrastive-loss2}
Florian Schroff, Dmitry Kalenichenko, and James Philbin.
\newblock Facenet: A unified embedding for face recognition and clustering.
\newblock \emph{2015 IEEE Conference on Computer Vision and Pattern Recognition
  (CVPR)}, pages 815--823, 2015.

\bibitem[Segal et~al.(2009)Segal, Hähnel, and Thrun]{generalized-icp}
Aleksandr Segal, Dirk Hähnel, and Sebastian Thrun.
\newblock Generalized-icp.
\newblock 06 2009.
\newblock \doi{10.15607/RSS.2009.V.021}.

\bibitem[Simon et~al.(2020)Simon, Fischer, Milz, Witt, and
  Gro{\ss}]{sticky-pillars}
Martin Simon, Kai Fischer, Stefan Milz, Christian Witt, and Horst-Michael
  Gro{\ss}.
\newblock Stickypillars: Robust feature matching on point clouds using graph
  neural networks.
\newblock \emph{ArXiv}, abs/2002.03983, 2020.

\bibitem[Sipiran and Bustos(2011)]{harris-3d}
Ivan Sipiran and Benjamin Bustos.
\newblock Harris 3d: A robust extension of the harris operator for interest
  point detection on 3d meshes.
\newblock \emph{The Visual Computer}, 27:\penalty0 963--976, 11 2011.
\newblock \doi{10.1007/s00371-011-0610-y}.

\bibitem[Steder et~al.()Steder, Rusu, KurtKonolige, and
  Burgard]{narf-3d-range-features}
Bastian Steder, Radu~Bogdan Rusu, KurtKonolige, and Wolfram Burgard.
\newblock Narf: 3d range image features for object recognition.

\bibitem[Sun et~al.(2009)Sun, Ovsjanikov, and
  Guibas]{a-concise-and-provably-informative-heat-diffusion}
Jian Sun, Maks Ovsjanikov, and Leonidas Guibas.
\newblock A concise and provably informative multi-scale signature based on
  heat diffusion.
\newblock In \emph{Proceedings of the Symposium on Geometry Processing}, SGP
  '09, page 1383–1392, Goslar, DEU, 2009. Eurographics Association.

\bibitem[Theiler et~al.(2014)Theiler, Wegner, and
  Schindler]{keypoint-based-4-pcs}
Pascal~Willy Theiler, Jan~Dirk Wegner, and Konrad Schindler.
\newblock Keypoint-based 4-points congruent sets – automated marker-less
  registration of laser scans.
\newblock \emph{ISPRS Journal of Photogrammetry and Remote Sensing},
  96:\penalty0 149--163, 2014.
\newblock ISSN 0924-2716.
\newblock \doi{https://doi.org/10.1016/j.isprsjprs.2014.06.015}.
\newblock URL
  \url{https://www.sciencedirect.com/science/article/pii/S0924271614001701}.

\bibitem[Tombari et~al.(2010{\natexlab{a}})Tombari, Salti, and Stefano]{SHOT}
F.~Tombari, S.~Salti, and L.~Di Stefano.
\newblock Unique signatures of histograms for local surface description.
\newblock \emph{Proc. ECCV}, 6313:\penalty0 356--369, 09 2010{\natexlab{a}}.
\newblock \doi{10.1007/978-3-642-15558-1_26}.

\bibitem[Tombari et~al.(2010{\natexlab{b}})Tombari, Salti, and
  Di~Stefano]{unique-shape-context-for-3d-data-description}
Federico Tombari, Samuele Salti, and Luigi Di~Stefano.
\newblock Unique shape context for 3d data description.
\newblock In \emph{Proceedings of the ACM Workshop on 3D Object Retrieval},
  3DOR '10, page 57–62, New York, NY, USA, 2010{\natexlab{b}}. Association
  for Computing Machinery.
\newblock ISBN 9781450301602.
\newblock \doi{10.1145/1877808.1877821}.
\newblock URL \url{https://doi.org/10.1145/1877808.1877821}.

\bibitem[Tombari et~al.(2010{\natexlab{c}})Tombari, Salti, and
  Di~Stefano]{unique-signatures-of-shistograms}
Federico Tombari, Samuele Salti, and Luigi Di~Stefano.
\newblock Unique signatures of histograms for local surface description.
\newblock In \emph{Proceedings of the 11th European Conference on Computer
  Vision Conference on Computer Vision: Part III}, ECCV'10, page 356–369,
  Berlin, Heidelberg, 2010{\natexlab{c}}. Springer-Verlag.
\newblock ISBN 364215557X.

\bibitem[Tran et~al.(2014)Tran, Bourdev, Fergus, Torresani, and
  Paluri]{c3d-3d-convolutional-networks}
Du~Tran, Lubomir~D. Bourdev, Rob Fergus, Lorenzo Torresani, and Manohar Paluri.
\newblock C3d: Generic features for video analysis.
\newblock \emph{ArXiv}, abs/1412.0767, 2014.

\bibitem[Wang et~al.(2022)Wang, Liu, Dong, Wang, and Yang]{YOHO}
Haiping Wang, Yuan Liu, Zhen Dong, Wenping Wang, and Bisheng Yang.
\newblock You only hypothesize once: Point cloud registration with
  rotation-equivariant descriptors.
\newblock \emph{ACM Multimedia 2022}, 2022.

\bibitem[Wang and
  Solomon(2019{\natexlab{a}})]{dcp-learning-representation-for-registration}
Yue Wang and Justin~M. Solomon.
\newblock Deep closest point: Learning representations for point cloud
  registration.
\newblock \emph{2019 IEEE/CVF International Conference on Computer Vision
  (ICCV)}, pages 3522--3531, 2019{\natexlab{a}}.

\bibitem[Wang and
  Solomon(2019{\natexlab{b}})]{prnet-self-supervised--for-partial-to-partial-registration}
Yue Wang and Justin~M. Solomon.
\newblock Prnet: Self-supervised learning for partial-to-partial registration.
\newblock In \emph{NeurIPS}, 2019{\natexlab{b}}.

\bibitem[Wu et~al.(2015)Wu, Song, Khosla, Yu, Zhang, Tang, and
  Xiao]{modelnet40}
Zhirong Wu, S.~Song, A.~Khosla, Fisher Yu, Linguang Zhang, Xiaoou Tang, and
  J.~Xiao.
\newblock 3d shapenets: A deep representation for volumetric shapes.
\newblock In \emph{2015 IEEE Conference on Computer Vision and Pattern
  Recognition (CVPR)}, pages 1912--1920, Los Alamitos, CA, USA, jun 2015. IEEE
  Computer Society.
\newblock \doi{10.1109/CVPR.2015.7298801}.
\newblock URL
  \url{https://doi.ieeecomputersociety.org/10.1109/CVPR.2015.7298801}.

\bibitem[Xu et~al.(2021)Xu, Liu, Wang, Liu, and
  Zeng]{omnet-learning-overlapping-mask}
Hao Xu, Shuaicheng Liu, Guangfu Wang, Guanghui Liu, and Bing Zeng.
\newblock Omnet: Learning overlapping mask for partial-to-partial point cloud
  registration.
\newblock \emph{2021 IEEE/CVF International Conference on Computer Vision
  (ICCV)}, pages 3112--3121, 2021.

\bibitem[Yang et~al.(2019)Yang, Xiao, and Cao]{3Dreconstruction_reference1}
J.~Yang, Y.~Xiao, and Z.~Cao.
\newblock Aligning 2.5d scene fragments with distinctive local geometric
  features and voting-based correspondences.
\newblock \emph{IEEE Transactions on Circuits and Systems for Video
  Technology}, 29\penalty0 (3):\penalty0 714--729, 2019.

\bibitem[Yang et~al.(2013)Yang, Li, and Jia]{go-icp}
Jiaolong Yang, Hongdong Li, and Yunde Jia.
\newblock Go-{ICP}: Solving 3{D} {R}egistration {E}fficiently and {G}lobally
  {O}ptimally.
\newblock In \emph{2013 IEEE International Conference on Computer Vision},
  pages 1457--1464, 2013.
\newblock \doi{10.1109/ICCV.2013.184}.

\bibitem[Yew and Lee(2018)]{3dfeat-net-weakly-supervised}
Zi~Jian Yew and Gim~Hee Lee.
\newblock 3dfeat-net: Weakly supervised local 3d features for point cloud
  registration.
\newblock \emph{ArXiv}, abs/1807.09413, 2018.

\bibitem[Yew and Lee(2020)]{rpm-net-robust-matching-using-learned-features}
Zi~Jian Yew and Gim~Hee Lee.
\newblock R{PM}-{N}et: Robust {P}oint {M}atching {U}sing {L}earned {F}eatures.
\newblock \emph{2020 IEEE/CVF Conference on Computer Vision and Pattern
  Recognition (CVPR)}, pages 11821--11830, 2020.

\bibitem[Yew and Lee(2022)]{REGTR}
Zi~Jian Yew and Gim~hee Lee.
\newblock Regtr: End-to-end point cloud correspondences with transformers.
\newblock In \emph{CVPR}, 2022.

\bibitem[Yu et~al.(2021)Yu, Li, Saleh, Busam, and
  Ilic]{cofinet-reliable-coarse-to-fine-correspondences}
Hao Yu, Fu~Li, Mahdi Saleh, Benjamin Busam, and Slobodan Ilic.
\newblock Cofinet: Reliable coarse-to-fine correspondences for robust point
  cloud registration.
\newblock In \emph{NeurIPS}, 2021.

\bibitem[Yuan et~al.(2020)Yuan, Eckart, Kim, Jampani, Fox, and Kautz]{deep-gmr}
Wentao Yuan, Benjamin Eckart, Kihwan Kim, V.~Jampani, Dieter Fox, and Jan
  Kautz.
\newblock Deep{GMR}: Learning {L}atent {G}aussian {M}ixture {M}odels for
  {R}egistration.
\newblock In \emph{ECCV}, 2020.

\bibitem[Yue et~al.(2018)Yue, Bisheng, Fuxun, and
  Zhen]{iterative-global-similarity-points}
Pan Yue, Yang Bisheng, Liang Fuxun, and Dong Zhen.
\newblock Iterative global similarity points: A robust coarse-to-fine
  integration solution for pairwise 3d point cloud registration.
\newblock In \emph{2018 International Conference on 3D Vision (3DV)}, 2018.

\bibitem[Zeng et~al.(2017)Zeng, Song, Nie{\ss}ner, Fisher, Xiao, and
  Funkhouser]{3dmatch}
Andy Zeng, Shuran Song, Matthias Nie{\ss}ner, Matthew Fisher, Jianxiong Xiao,
  and Thomas Funkhouser.
\newblock 3dmatch: Learning local geometric descriptors from rgb-d
  reconstructions.
\newblock In \emph{CVPR}, 2017.

\bibitem[Zhong(2009)]{intrinsic-shape-signatures}
Yu~Zhong.
\newblock Intrinsic shape signatures: A shape descriptor for 3d object
  recognition.
\newblock In \emph{2009 IEEE 12th International Conference on Computer Vision
  Workshops, ICCV Workshops}, pages 689--696, 2009.
\newblock \doi{10.1109/ICCVW.2009.5457637}.

\end{thebibliography}
